\documentclass[runningheads]{llncs}

\usepackage[utf8]{inputenc}
\usepackage{graphicx}
\usepackage{xspace}
\usepackage{listings}
\usepackage{multirow}
\usepackage{amsmath}
\usepackage{todonotes}
\usepackage{cite}
\usepackage{amsfonts}

\usepackage{hyperref}
\usepackage[nameinlink]{cleveref}

\usepackage{inconsolata} %
\title{Towards Portfolios of Streamlined Constraint Models: A Case Study with the Balanced Academic Curriculum Problem}
\titlerunning{Towards Portfolios of Streamlined Constraint Models}

\author{Patrick Spracklen, Nguyen Dang, Özgür Akgün, Ian Miguel}

\institute{School of Computer Science, University of St Andrews, St Andrews, UK\\
\{jlps, nttd, ozgur.akgun, ijm\}@st-andrews.ac.uk}

\authorrunning{Spracklen, Dang, Akgun, Miguel}

\begin{document}

\newcommand{\conjure}[0]{\textsc{Conjure}\xspace}
\newcommand{\compact}[0]{\textsc{compact}\xspace}
\newcommand{\default}[0]{\textsc{default}\xspace}
\newcommand{\sparse}[0]{\textsc{Sparse}\xspace}
\newcommand{\essence}[0]{\textsc{Essence}\xspace}
\newcommand{\eprime}[0]{\textsc{Essence Prime}\xspace}
\newcommand{\savilerow}[0]{\textsc{Savile Row}\xspace}
\newcommand{\chuffed}[0]{\textsc{Chuffed}\xspace}
\newcommand{\lingeling}[0]{\textsc{Lingeling}\xspace}
\newcommand{\tailor}[0]{\textsc{Tailor}\xspace}
\newcommand{\minion}[0]{\textsc{Minion}\xspace}
\newcommand{\dominion}[0]{\textsc{Dominion}\xspace}
\newcommand{\flatzinc}{FlatZinc\xspace}
\newcommand{\irace}[0]{\textsc{irace}\xspace}
\newcommand{\athanor}[0]{\textsc{Athanor}\xspace}
\newcommand{\code}[1]{{\small\texttt{#1}}}
\newcommand{\comment}[1]{\textcolor{blue}{#1}}

\newcommand{\oz}[1]{\todo[backgroundcolor=green!20!white]{#1}}
\newcommand{\nguyen}[1]{\todo[backgroundcolor=blue!20!white]{#1}}
\newcommand{\patrick}[1]{\todo[backgroundcolor=purple!20!white]{#1}}

\maketitle

\begin{abstract}
Augmenting a base constraint model with additional constraints can strengthen the inferences made by a solver and therefore reduce search effort. We focus on the automatic addition of streamliner constraints, derived from the types present in an abstract \essence specification of a problem class of interest, which trade completeness for potentially very significant reduction in search. The refinement of streamlined \essence specifications into constraint models suitable for input to constraint solvers gives rise to a large number of modelling choices in addition to those required for the base \essence specification. Previous automated streamlining approaches have been limited in evaluating only a single default model for each streamlined specification. In this paper we explore the effect of model selection in the context of streamlined specifications. We propose a new best-first search method that generates a portfolio of Pareto Optimal streamliner-model combinations by evaluating for each streamliner a portfolio of models to search and explore the variability in performance and find the optimal model. Various forms of racing are utilised to constrain the computational cost of training. 
We demonstrate the approach's effectiveness on the Balanced Academic Curriculum Problem, which involves assigning periods to courses in a way that the academic load of each period is balanced.

\end{abstract}

\section{Introduction}

Adding {\em streamliner} constraints \cite{gomes2004streamlined}  to a model offers a powerful means to improve the performance of a basic constraint model. Streamliners differ from implied constraints \cite{charnley2006automatic,colton2001constraint,frisch2003cgrass}, which are inferred from the original model, and symmetry-breaking \cite{frisch2007symmetry} and dominance-breaking constraints \cite{prestwich2004exploiting}, both of which rule out members of equivalence classes of solutions. Streamliners are conjectured rather than inferred from the original model and are hence not guaranteed to be sound. As a result they can very significantly alter the set of solutions to a problem instance. A well chosen streamliner can substantially reduce search effort by focusing search on promising regions of the search space containing at least one solution.

Streamliner generation was originally a laborious manual process, requiring the manual inspection of small instances of a problem class to identify patterns from which to derive candidate streamliners
\cite{kouril2005resolution,lebras2013double}. This approach was successful in several domains, such as generating spatially-balanced experimental designs \cite{gomes2004streamlined} and finding improved bounds for the Erdos discrepancy problem \cite{le2014erdHos}.

More recently, it has been shown how this process can be automated to derive candidate streamliners directly from the \essence specification of a problem class of interest \cite{spracklen2018automatic,spracklen2019automatic}. \essence \cite{frisch2005essence,frisch2007design,frisch2008ssence} is an abstract constraint specification language that allows a problem to be described directly in terms of the combinatorial structure to be found, through support for abstract type constructors like set, relation, function and partition, as well as arbitrary nesting of these, such as set of multisets. 
Figure ~\ref{fig:Problems_BACP} presents the \essence specification of the Balanced Academic Curriculum Problem (BACP), which we will use throughout. Here, the problem involves finding a total function from courses to periods.

The structure apparent in an \essence specification drives a rule-based system to suggest candidate streamliners, followed by a best-first search to identify the most promising streamliners to apply in practice. For example one of the streamliners automatically generated for BACP enforces that the \emph{Courses} can only be placed into \emph{even} numbered \emph{Periods}. 

\begin{figure}[t!]
\centering
\begin{lstlisting}[breaklines,basicstyle=\scriptsize]
given n_courses, n_periods,load_per_period_lb, load_per_period_ub,
      courses_per_period_lb, courses_per_period_ub : int(1..)

letting Course be domain int(1..n_courses),
        Period be domain int(1..n_periods)
given prerequisite : relation of (Course * Course),
      course_load : function (total) Course --> int(1..)

find curr : function (total) Course --> Period

such that
    forAll c1,c2 : Course . prerequisite(c1,c2) -> curr(c1) < curr(c2),
    forAll p : Period . 
      (sum c in preImage(curr,p). course_load(c)) <= load_per_period_ub
        /\
      (sum c in preImage(curr,p). course_load(c)) >= load_per_period_lb,
    forAll p : Period .
      |preImage(curr,p)| <= courses_per_period_ub /\ 
      |preImage(curr,p)| >= courses_per_period_lb

\end{lstlisting}
\caption{\essence{} specification of the Balanced Academic Curriculum Problem~\cite{csplib:prob030}. The goal is to design a balanced academic curriculum by assigning periods to courses in a way that the academic load of each period is balanced, i.e., as similar as possible.}
\label{fig:Problems_BACP}
\end{figure}

The existing work~\cite{spracklen2018automatic,spracklen2019automatic} uses the \conjure automated modelling system~\cite{akgun2011extensible,akgun2013automated} to refine a streamlined \essence specification into a constraint model for evaluation. This work is limited to accepting \conjure's default choice for each of the modelling decisions that need to be made, such as how to represent abstract variables as constrained collections of more primitive variables, resulting in a single streamlined model. The hypothesis that motivates the work in this paper is that these default modelling choices are unlikely always to result in the most effective model, and that the best modelling choices may vary according to which streamliner constraints are added to the original specification.

To test this hypothesis, we use \conjure to refine each streamlined specification to a set of candidate models, each representing a different set of modelling choices. We augment the best-first search to look for effective streamliner/model pairs to build a portfolio of effective streamlined models for the problem class under consideration. For an unseen instance from the problem class, a model from the portfolio is automatically selected and deployed. For the BACP, our empirical results demonstrate that this often results in a significant improvement in performance over the existing method of focusing on a single default set of modelling choices.

\section{Candidate Streamliner Generation}

An \essence{} specification like the one presented in Figure~\ref{fig:Problems_BACP} comprises the problem class parameters ({\tt given}); the combinatorial objects to be found ({\tt find}); the constraints the objects must satisfy ({\tt such that}); identifiers declared ({\tt letting}); and an optional objective function ({\tt min/maximising}).

The highly structured problem description an \essence{} specification provides is better suited to streamliner generation than a lower level representation, such as MiniZinc \cite{nethercote2007minizinc} or {\sc Essence Prime} \cite{nightingale2016essence}. This is because abstract types like function or partition and nested types like multiset of sets must be represented as a constrained collection of more primitive variables in a constraint model, obscuring the structure used to drive streamliner generation.

We employ the same set of streamliner generation rules as \cite{spracklen2018automatic,spracklen2019automatic}. These rules are generic, developed around the structured abstract types provided by \essence. The subset that are applicable to the BACP is summarised in \Cref{table:rules}. Higher-order rules take another rule as an argument and lift its operation onto a decision variable with a nested domain such as the function present in the BACP. This allows for the generation of a rule that enforces that half of the \emph{Courses} are assigned to \emph{Periods} drawn from the lower-half of the \emph{Periods} domain.

\begin{table}
  \centering
  \begin{tabular}{|c|c|c|}
    \hline
    Class & Trigger Domain & Name \\
    \hline
    \multirow{2}{*}{First-order}
      & \multirow{1}{*}{\tt int}
      &  odd\{even\}, lower\{upper\}Half\\
    \cline{2-3}
    & \multirow{2}{*}{{\tt function int --> int}}
      & monotonicIncreasing\{Decreasing\}\\
      & & largest\{smallest\}First\{Last\} \\
    \hline
    \multirow{2}{*}{Higher-order}
    & \multirow{1}{*}{\tt set of X}
    &  all, most$^*$, half, approxHalf$^*$\\
    \cline{2-3}
    & \multirow{2}{*}{\tt function X --> Y}
    &  range, defined \\
    & &  pre\{post\}fix$^*$, allBut$^*$\\
        \cline{2-3}
    \hline
  \end{tabular}
  \vspace{3mm}
  \caption{The subset of rules used to generate streamliners for BACP. The ones with a softness parameter specify a family of rules each member of which is defined by an integer parameter, and are marked with $^*$. Note that the defined and range operators return sets, and the higher order rules that operate on sets are applicable to these sets.}
  \label{table:rules}
\end{table}

\section{From \essence Specifications to Constraint Models}

\essence specifications cannot be solved directly by systematic constraint solvers, which lack support for abstract types such as functions and partitions, and nested types. 
Hence, an \essence specification must be {\em refined} into a constraint model where these types, and the constraints on them, are modelled in terms of the target constraint language. The automated modelling tool \conjure{} \cite{akgun2014extensible,akgun2014breaking} refines \essence specifications into constraint models in the solver-independent constraint language \eprime{} \cite{nightingale2016essence, nightingale2017automatically}. 

There are typically many alternative models of an \essence specification, corresponding to alternative modelling choices for the abstract types, and the constraints upon them, present in the specification. Different models can exhibit significantly different performance behaviour, which can also vary according to the particular instance of the problem class modelled. \conjure is able to enumerate models of an \essence specification by applying the different refinement rules corresponding to the possible modelling choices.

We use the BACP (\Cref{fig:Problems_BACP}) to illustrate how \conjure{} produces alternative models. This specification has one \emph{find} statement with a function domain: \emph{curr}. This can be modelled in four different ways in the current version of \conjure. As this is a \emph{total} function the first representation uses a 1-D matrix \texttt{matrix indexed by [int(1..n\_courses)] of int(1..n\_periods)}. The second model represents the function as a \emph{total} relation. It utilises a 2-D matrix of Booleans \texttt{matrix indexed by [int(1..n\_courses), int(1..n\_periods)] of bool}. An alternative interpretation models the function as a binary relation first, followed by modelling the relation in two different ways. Both representations use two integer matrices \texttt{matrix indexed by [int(1..n\_courses * n\_periods)] of int(1..n\_courses)]} for each component of the relation. They differ in how they represent the cardinality of the function, one using a single integer marker and the other one using a Boolean matrix as flags to denote whether the value is in the relation. 

\begin{figure}
    \centering
    \includegraphics[width=\textwidth]{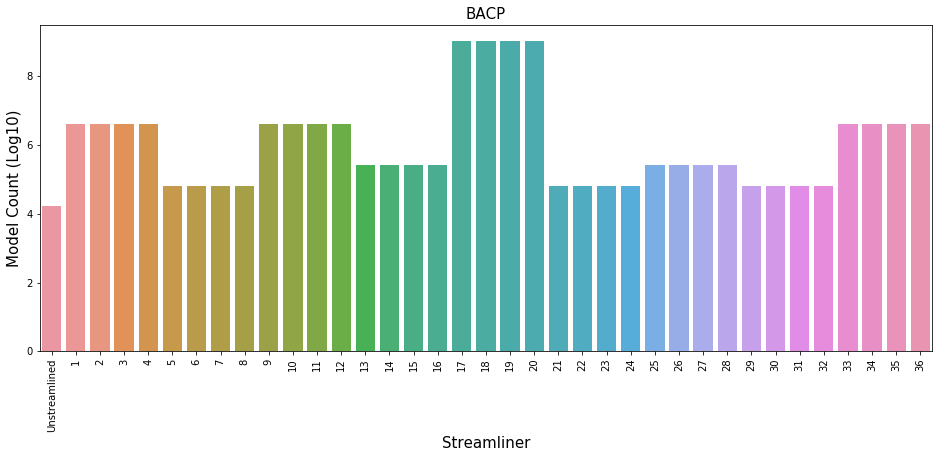}
    \caption{The number of models ($log_{10}$) refined from the original \essence specification and with a sample set of single candidate streamliners for BACP. The streamliners are represented in \conjure{} via numeric values which are presented.}
\label{fig:model_count}
\end{figure}

From this one \emph{find} statement there are four possible base representations. However, \conjure{} also has to model other types of objects such as \emph{given} parameters, \emph{auxiliary} variables and \emph{quantified} variables each of which could have multiple different representations. \conjure{} also has the ability to implement \emph{automated channelling} \cite{cheng1999increasing} in which each reference to a variable in a constraint expression can be modelled differently.
Each of these components increases the number of modelling choices and as a result a single abstract \essence{} specification can typically be refined into a large number of constraint models. \Cref{fig:model_count} shows for the number of candidate models of BACP on the unstreamlined \essence{} and how this changes with the introduction of streamliner constraints. Over 16,384 different models can be generated for the unstreamlined specification alone and with the additional modelling choices streamliners radically increase this number. Even though a large number of these models are small modifications of other models, some models are significantly different from others and they have very different performance characteristics. Moreover, different models are also likely to benefit from the addition of streamliner constraints differently.

In existing work \cite{spracklen2018automatic, spracklen2019automatic} this choice is handled by utilising the \default{} heuristic of \conjure{}. This is a heuristic employed during refinement to commit greedily to promising modelling choices at each point where an abstract type or a constraint expression may be refined in multiple ways~\cite{akgun2013automated}. The \default{} heuristic is a combination of two other heuristics: \compact{} for decision variables and constraint expressions and \sparse{} for parameters. \compact{} favours transformations that produce smaller expressions. For an abstract type, we define an ordering as follows: concrete domains (such as bool, matrix) are smaller than abstract domains; within concrete domains, bool is smaller than int and int is smaller than matrix. Abstract type constructors have the ordering set $<$ multiset (mset) $<$ function $<$ relation $<$ partition. These rules are applied recursively so that \compact{} will select the smallest domain according to this order. The \sparse{} heuristic is designed to choose the domain representations that will generate the smallest \eprime{} parameter files for sparse objects. Some representations represent every \textit{potential} member of a domain explicitly whereas some representations only represent the actual members of a domain. 

\begin{figure}
    \centering
    \includegraphics[width=\textwidth]{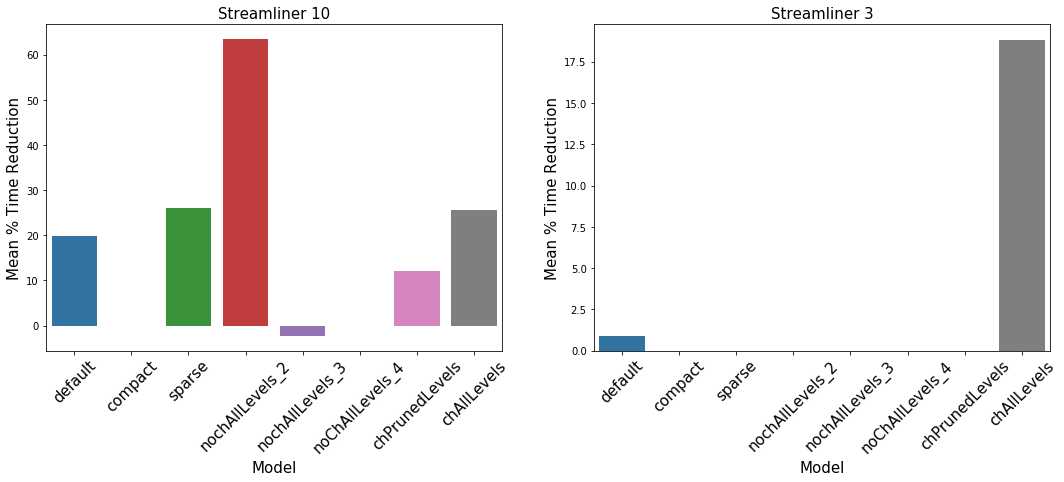}
    \caption{Performance of two example streamliners across 8 models for BACP. Each model is named with respect to the generating heuristic (\cref{table:results}) }
    \label{fig:model_streamliner_comparison}
\end{figure}

\Cref{fig:model_streamliner_comparison} shows the performance of streamlined models for BACP across 8 different models with two example streamliners. Streamliner 3 enforces that the smallest integer courses must be assigned to the smallest integer periods, while Streamliner 10 enforces that half of the courses must be placed in even numbered periods. 
Results in~\Cref{fig:model_streamliner_comparison} indicates that  there can be a large variability in the streamliner performance across different model representations. Moreover,  the \default{} heuristic may not always provide the best performance, as for streamliner 10 the \default{} model is only able to achieve around a 20\% reduction in time, far inferior to the 60\% reduction achieved by the best model. It is also interesting to note that the best model representation is highly streamliner dependent and changes between streamliner 10 and streamliner 3.

\section{Model Portfolios}

As seen from \Cref{fig:model_count} the number of possible models that can be refined, especially from a streamlined specification, means that refining and evaluating all possible models for each streamliner is not feasible. Instead, a sample of $N$ models can be evaluated where $N$ is controlled to balance the computational cost against the exploratory benefit. \conjure{} allows for the customisation of the modelling process in several ways. First, different strategies can be employed for the modelling of different types of objects. A distinction can be made among four different forms of declarations, \emph{\{find, given, auxiliary, quantified\}} depending upon their origin in the specification. \textit{find} and \textit{given} statements define decision variables and problem parameters, \textit{auxiliaries} are decision variables created by \conjure{} during model reformulation, and \textit{quantified variables} are defined by quantified expressions like \texttt{forAll}, \texttt{exists} and \texttt{sum}. Second, automated channelling and precedence levels can be activated and deactivated to further customise the modelling heuristic. To produce a portfolio of $N$ models we use a set of heuristics, as listed in \Cref{table:results}, ranked based on the predicted number of generated models and our perceived effectiveness of the generated models. These heuristics are then invoked in an iterative fashion until a portfolio of $N$ models has been generated. 

\begin{table}
\let\center\empty
\let\endcenter\relax
\centering
\resizebox{.8\width}{!}{\begin{tabular}{|c|l|l|l|l|l|l|l|}
\hline
\multirow{2}{*}{\textbf{Rank}} & \multirow{2}{*}{\textbf{Heuristic}} & \multicolumn{4}{c|}{\textbf{Declaration Type}}                                                  & \multicolumn{2}{c|}{\textbf{Additional Flags}}                             \\ \cline{3-8}
                                                     & & \textbf{Find} & \textbf{Given} & \textbf{Auxiliaries} & \textbf{Quantifieds} & \textbf{Channelling} & \textbf{Levels}                \\ \hline
 1 & {\sc Default}                                           & {\sc Compact}       & {\sc Sparse}         & {\sc Compact}              & {\sc Compact}                  & False                & True                          \\ \hline
 2 & {\sc Compact}                                           & {\sc Compact}       & {\sc Compact}        & {\sc Compact}              & {\sc Compact}                  & False                & True                          \\ \hline
 3 & {\sc Sparse}                                            & {\sc Sparse}        & {\sc Sparse}         & {\sc Sparse}               & {\sc Sparse}                   & False                & True                          \\ \hline
 4 & nochPrunedLevels                                  & All           & {\sc Sparse}         & {\sc Compact}              & {\sc Compact}                  & False                & True                          \\ \hline
 5 & nochAllLevels                                     & All           & {\sc Sparse}         & {\sc Compact}              & {\sc Compact}                  & False                & False                         \\ \hline
 6 & chPrunedLevels                                    & All           & {\sc Sparse}         & {\sc Compact}              & {\sc Compact}                  & True                 & True                          \\ \hline
 7 & chAllLevels                                       & All           & {\sc Sparse}         & {\sc Compact}              & {\sc Compact}                  & True                 & False                         \\ \hline
 8 & fullPrunedLevels                                  & All           & {\sc Sparse}         & All                  & All                      & True                 & True                          \\ \hline
 9 & fullAllLevels                                     & All           & {\sc Sparse}         & All                  & All                      & True                 & False                         \\ \hline
10 & fullParamsPrunedLevels                            & All           & All            & All                  & All                      & True                 & True                          \\ \hline
11 & fullParamsAllLevels                               & All           & All            & All                  & All                      & True                 & False                         \\ \hline
\end{tabular}
}
\caption{The ranked set of heuristics we use when generating a portfolio of models.}
\label{table:results}
\end{table}

\section{Searching for Streamlined Model Portfolios}\label{sec:searching}

\subsection{Multi-Objective Monte Carlo Tree Search}
Candidate streamliners are often most effectively used in combination \cite{gomes2004streamlined}, forming a lattice of possible combinations \cite{wetter2015automatically}, and therefore a search allowing good combinations to be identified efficiently is desirable. In reality, streamliner generation has two conflicting goals: to uncover constraints that steer search towards a small and highly structured area of the search space that yields a solution, versus identifying streamliner constraints in training that remain applicable to more difficult instances of the problem class.

To address these two objectives we employ a multi-objective optimisation approach, where each point $x$ in the search space $X$ is associated with a $2$-dimensional reward vector $r_x$ in $\mathbb{R}^2$. Our two objectives allow us explicitly to balance considerations of average applicability against how aggressively the streamlined version of each candidate model reduces search: 
(i) \textit{Applicability} - the proportion of training instances for which the streamlined model admits a solution. (ii) \textit{Search Reduction} - the mean reduction in time to find a solution in comparison with an unstreamlined model.
All objectives are transformed such that they can be maximised. With these two objectives for each streamliner combination, the Pareto dominance partial ordering $\succ$ on $\mathbb{R}^2$ is defined as follows. 
Given $x, y \in X$ with vectorial rewards $r_x = \langle r^1_x, r^2_x\rangle$ and $r_y = \langle r^1_y, r^2_y \rangle$: 

\begin{equation}
r_x \succ r_y 
\iff (\forall i \in [1, 2], r^i_x \geq r^i_y)
\land (\exists j \in [1, 2],  r^j_x > r^j_y)
\end{equation}

To search the lattice structure for a portfolio of Pareto optimal streamlined models we have adapted the \emph{Dominance-based Multi-Objective Monte Carlo Tree Search (MOMCTS-DOM)} algorithm \cite{wang2013hypervolume}. The algorithm has four phases:

\begin{enumerate}
\item \textbf{Selection}: Starting at the root node, the Upper Confidence Bound applied to Trees
(UCT)  \cite{Browne12asurvey} policy is applied to traverse the explored part of the lattice until an unexpanded node is reached.%
\item \textbf{Expansion}: Uniformly select and expand an admissible child.
\item \textbf{Simulation}: For each of a portfolio of $N$ models the collection of streamliners associated with the expanded node are evaluated. The vectorial reward $\langle$Applicablity, Search Reduction$\rangle$ across the set of training instances is calculated and returned. 
\item \textbf{BackPropagation}: The current portfolio, which contains the set of non dominated streamliner/model combinations found up to this point during search, is used to compute Pareto dominance. The reward values of the Pareto dominance test are non stationary since they depend on the portfolio, which evolves during search. Hence, we use the cumulative discounted dominance (CDD) \cite{wang2013hypervolume} reward mechanism during reward update. 

\end{enumerate}

\subsection{Training Set Construction}
It is necessary to have a training set with which to evaluate the streamliner candidates. We use the automated instance generation system proposed in~\cite{akgun2019instance} to generate several satisfiable instances, solvable in a time limit of $[10s, 300s]$. Starting from an Essence specification of a problem class, instances can be generated in a completely automated fashion. Ideally the training set should be diverse and representative of the problem class as a whole, otherwise the streamliner portfolio may not generalise. Hence, three different instance generation runs per problem class are performed to ensure diversity. Instance-specific features are used to judge instance similarity and GMeans clustering~\cite{hamerly2004learning} is employed to detect the number of instance clusters present.
Instances are then selected per cluster to build a representative training set.
For instance features, we use the $95$ \flatzinc features provided by the \code{fzn2feat} tool (part of \code{mzn2feat}~\cite{Amadini2014:enhanced}).

\subsection{Multi-Round Search}
One issue with the search method presented thus far is that because of the domination criteria used the resultant portfolio may still be sub-optimal. A streamliner/model combination that has poor average performance across all instances it covers may still be useful, as it might provide very high solving-time reduction within a subset of those instances. However, this information is not considered in the current dominance criteria, and will be discarded.

In order to solve this issue, we adapt our search to incorporate the ideas of Hydra \cite{xu2010hydra}, a portfolio builder approach that automatically builds a set of solvers or parameter configurations of solvers with complementary strengths. Instead of performing one lattice search and building one portfolio, we now perform multiple rounds of search. In the first, a normal MOMCTS-DOM is conducted. From the second round onward, the best solving time on each instance obtained from the portfolio created in previous rounds is recorded. Performance of a streamliner/model combination on an instance in the current round is then measured relative to the previously recorded best solving time on that instance. This mechanism allows the portfolio built in the current round to complement the strengths of the combined portfolios built in the prior rounds.

\section{Model Racing}

Evaluating all candidate models for each considered streamliner combination is expensive. Poorly-performing streamliner/model combinations can timeout on several instances and consume a large amount of the training budget. Hence, we employ various {\em racing} techniques~\cite{birattari2002racing,hutter2009paramils,akgun2013automated,caceres2017experimental} to terminate poorly-performing combinations early. The streamliner search can then allocate more time to evaluate promising streamliners and gain a more accurate estimation of their performance.

Racing is performed at each lattice node of the search, representing a combination of streamliners. We race among the multiple models generated by \conjure on the given streamliner combination, and try to eliminate the poorly performing models without having to fully evaluate them. We first describe the three racing strategies we used, {\em $\rho$-Capping}~\cite{akgun2013automated}, {\em racing based on statistical tests}~\cite{birattari2002racing,lopez2016irace} and {\em adaptive capping} \cite{hutter2009paramils} and then discuss how they are combined together in our streamliner search through a multi-level model generation approach.

\subsection{$\rho$-Capping}
We conduct a race across multiple models for each training instance. Given a parameter $\rho \geq 1$, a model is {\em $\rho$-dominated}~\cite{akgun2013automated} on an instance by another model if the solving time of the latter is at least $\rho$ times faster on the given instance. All models enter each race, but a model is terminated as soon as it is $\rho-$dominated by some other model. 
Hence, the running time of a model on an instance is {\em capped} by $\rho$ times the best solving time of all models in the race. In our experiments $\rho$ is set as 2.

\subsection{Racing using statistical tests}
Racing using statistical tests for the automated configuration of parameterised algorithms~\cite{hoos2011automated} was first proposed in~\cite{birattari2002racing}, and later extended and implemented in the automated algorithm configuration tool \irace~\cite{lopez2016irace}. Bad algorithm configurations are discarded from a race as soon as sufficient statistical evidence is observed. We apply the same idea to remove bad models early during a race.

At the beginning of a race, all models generated by \conjure at the current lattice node are evaluated on a number of instances, denoted by $T_{first}$, and their solving times are measured. The model with the best average solving time on those instances is identified, and a {\em paired Student t-test} (with a significance level of $0.05$) between that model and each of the other models is conducted. {The \em Bonferroni correction} is used for adjusting the multiple-comparison statistical tests. Models showing statistically significantly worse performance than the current best-average one are eliminated. The survivors  continue to be evaluated on an additional number of $T_{next}$ instances before a new best model is calculated and statistical tests are applied again. This is repeated until all training instances are examined. We use \irace's default values for $T_{first}$ ($10$) and $T_{next}$ ($5$).

\subsection{Adaptive Capping}

Adaptive capping is another technique in automated algorithm configuration to terminate poorly-performing algorithm configurations early when optimising running time of a parameterised algorithm. It was first proposed in the local search-based automated algorithm configuration tool ParamILS~\cite{hutter2009paramils}, and was later integrated into \irace~\cite{caceres2017experimental}. The technique has been shown to significantly speed up the search for the best algorithm configurations in many cases~\cite{hutter2009paramils,hutter2011sequential,caceres2017experimental}. Our streamliner search makes use of the adaptive capping mechanism of \irace.

The main idea is to reduce the time wasted in the evaluation of poorly performing models in the current race by enforcing an upper bound on their running time. Bounds are calculated based upon a set of {\em elite} models (best-performing models) obtained from a previous race. More specifically, let $M$ denote a set of candidate models in the current race, $I=\{i_1,i_2,..,i_n\}$ a set of instances, $M_E$ a set of elite models and $p_{k}^{m}$ the average solving time of a model $m$ on the instance subset $\{i_1,i_2,..,i_k\}$ ($k \leq n$). Given the fact that we already know the values of $p^{m_e}_{k}$ for all $m_e \in M_E$ and $k \leq n$ from the previous race, the maximum running time $b_{k+1}^{m}$ ($k<n$) for a model $m \in M$ on instance $i_{k+1}$ is calculated as: $median_{m_e \in M_E}\{p^{m_e}_{k+1}\} * (k+1) - p^{m}_{k}*k$. 
This bound can be thought of as the minimum performance a current model must achieve on an instance subset to be competitive with the elites up to that point.
If the running time of a model ever exceeds the calculated bound, it is eliminated from the race. 
As the bound for an instance is calculated based upon the time taken on preceding instances, adaptive capping can be sensitive to the ordering of the instance set $I$. Therefore, the training instance set is shuffled at the beginning of every race.

\subsection{Multi-Level Model Generation}

At each lattice node of the streamliner search, the three forms of racing are combined in a multi-level model generation approach. On each level, a portfolio of models is generated and a new race is started. The size of the portfolio is doubled with each level and the model set created is a strict superset of the models in any preceding level. On the first level a single model is generated (\default model) and is evaluated on the whole training set. 
Since it is currently the only model, it becomes an elite for the subsequent level. The second level is then started with a portfolio of size 2, containing \default{} and a newly created model. On this level all three racing techniques are utilized to eliminate poorly performing models. 
Adaptive capping will make use of the elite models from the preceding level to calculate its bound and perform its elimination. Any results from previous levels that have already been calculated are cached and reused.
In our experiments, we allow a maximum of four levels, which results in a maximum number of 8 streamlined models at each lattice node.

Iterative deepening of the portfolio size allows us to initially evaluate the models that we believe are most likely to perform best based upon the ranking in \Cref{table:results}. The use of elite models derived from previous levels helps the adaptive capping mechanism to speed up the evaluations in the next levels substantially. This allows the search to explore larger portfolio sizes within a reasonable amount of time. It is especially useful where the truly best model for a particular streamliner combination is actually located in a lower ranked heuristic.

\section{Streamliner/Model Selection} \label{sec:algorithm_scheduling}
The generated portfolio for a problem class contains streamliner/model combinations with complementary strengths. For an unseen instance the question arises as to which streamlined model should be used. This is a common situation in many AI domains. Algorithm Selection~\cite{rice1976algorithm} uses instance characteristics to select from a set of algorithms the ones expected to solve a given problem instance most efficiently, and has been shown empirically to improve the state-of-the-art in several domains~\cite{kotthoff2016algorithm,kerschke2019automated}. We exploit existing Algorithm Selection techniques to effectively select from our streamliner portfolios for unseen test instances.

We use Autofolio~\cite{lindauer2015autofolio}, an automatically self-configured algorithm selection framework. Given a particular problem instance the goal is to have Autofolio predict, based upon the features of the instance, which streamlined models from the generated portfolio will most efficiently solve the instance. For instance features, we again use the \flatzinc features provided by the \code{fzn2feat} tool~\cite{Amadini2014:enhanced}. 
Autofolio also supports a \emph{pre-solving schedule}, a static schedule built from a small subset of streamlined models. This schedule is run for a small amount of time. If it fails to solve an instance, the model chosen by the prediction model is applied. Autofolio chooses whether to use a pre-solving schedule during its tuning phase. For each problem class, Autofolio was run ten times in tuning mode with a one day CPU budget. 10-fold cross validation was used on the training sets and the model with the best average cross-validation score was used for prediction.

\section{Experimental Results}\label{sec:results}

Experiments were run on a cluster of 280 nodes, each with two 2.1 GHz, 18-core Intel Xeon E5-2695 processors.
MOMCTS-DOM was run with leaf parallelisation~\cite{cazenave2007parallelization} using up to 30 cores with a maximum budget of 4 wall-time days per problem class. Difficult test instances, with solving time in $[300s, 3600s]$, were generated automatically using the same instance generation system~\cite{akgun2019instance} used for training instance construction. We will show that although the portfolios of streamlined models are trained on easy instances (solved in $[10s,300s]$), their large improvement in solving time transfers to difficult, unseen test instances.

Previous automated streamlining approaches have been limited to using only a single \default{} model for each streamlined specification. To demonstrate the positive impact of adding multiple-model exploration during streamliner search, two independent searches were conducted for each problem class: one with the single default model, and one with the multiple-model approach proposed in this paper. The generated portfolio of streamlined models from each search was then given to Autofolio to build a streamlined-model selector using a set of training instances. Performance of each streamliner search is measured as the performance of the selector over the test instances. Both searches were given the same training time budget for a fair comparison. Our hypothesis is that even though each round of a multiple-model search is generally more expensive than its single-model counterpart, the opportunity of exploring multiple models and our enhanced search method using model-racing will be able to produce higher-quality portfolios of streamlined models within the same time budget. 

The CP learning solver \chuffed~\cite{Chu2012:exploiting} was used with its default parameter setting as the target solver. Each streamlined model was generated by \conjure and further tailored for the target solver by \savilerow~\cite{sr-journal-17,nightingale2015automatically}. \savilerow was also used to produce the FlatZinc inputs for the \texttt{fzn2feat} feature extraction tool.

The evaluation of a test instance proceeds as follows: Initially the features are extracted (by \texttt{fzn2feat}) and used by Autofolio to predict the best streamliner for that part of the instance space. The selected streamliner is then evaluated for a maximum time period of one hour. As streamliners are not necessarily sound and no Algorithm Selection system is perfect in its predictions there is no guarantee that the streamliner will find a solution and can either time out or be proven unsatisfiable by the solver. In either case after the streamliner is finished we revert back to running the original model until a solution is found. The overall time for a streamliner is then calculated as follows:

\[ Total Time =
   \begin{cases}
     \text{SAT:} &  \text{streamlined solving time} \\
     \text{UNSAT/timeout:} &  \text{streamlined} + \text{original model solving time} \\
   \end{cases}
\]

Table \ref{table:results_new} shows our experimental results of the single-model and multiple-model settings on the 25 difficult test instances generated. We report results of Autofolio and the Virtual Best Solver (VBS) for each setting using three measurements. {\em \%imp} and {\em \%red} represent the percentage of test instances where the solving time is improved by adding streamliners and the average solving-time reduction percentage on those improved instances respectively. {\em speedup} is calculated as the average solving time of the original model divided by the average streamlined \emph{TotalTime} across all test instances. 

\begin{table}
\centering
\let\endcenter\relax
\begin{tabular}{l|c|rrr|rrr}
\multirow{2}{*}{Experiment} & \multirow{2}{*}{\#inst} & \multicolumn{3}{c|}{VBS}                & \multicolumn{3}{c}{Autofolio}           \\
\multicolumn{1}{l|}{}                            &                         & \%imp         & \%red & speedup          & \%imp         & \%red & speedup           \\
\hline
BACP (D)                                        & 25                      & 92.0          & 95.06 & 8.31          & 72.0          & 96.0  & 1.23          \\
BACP (MM)                                       & 25                      & \textbf{96.0} & 98.7 & \textbf{14.6} & \textbf{92.0} & 92.0  & \textbf{3.6}  \\
\hline
\end{tabular}

\caption{Results of the Virtual Best Solver (VBS) and Autofolio on 25 test instances for BACP, under the \default{} model setting (D) and the multi-model (MM) setting. {\em \%imp} and {\em \%red} represent the percentage of test instances where the solving time is reduced by adding streamliners and the average solving-time reduction percentage on those improved instances respectively. {\em speedup} represents the relative performance between the average solving time of the unstreamlined model in comparison to the two streamlined approaches.  }
\label{table:results_new}
\end{table}

Both streamlined approaches provide large improvement in solving time to the original unstreamlined model. Compared to the single-model approach, the multi-model search results in higher percentages of instances with improvement in search time ({\em \%imp}) and higher speedup values. These results are achieved both on the VBS and with Autofolio which shows that the multi-model portfolios not only performing better in theory (VBS) but also that these theoretical gains can substantially be realised in practice by an Algorithm Selector (Autofolio).

\begin{figure}
    \centering
    \includegraphics[width=0.4\textwidth]{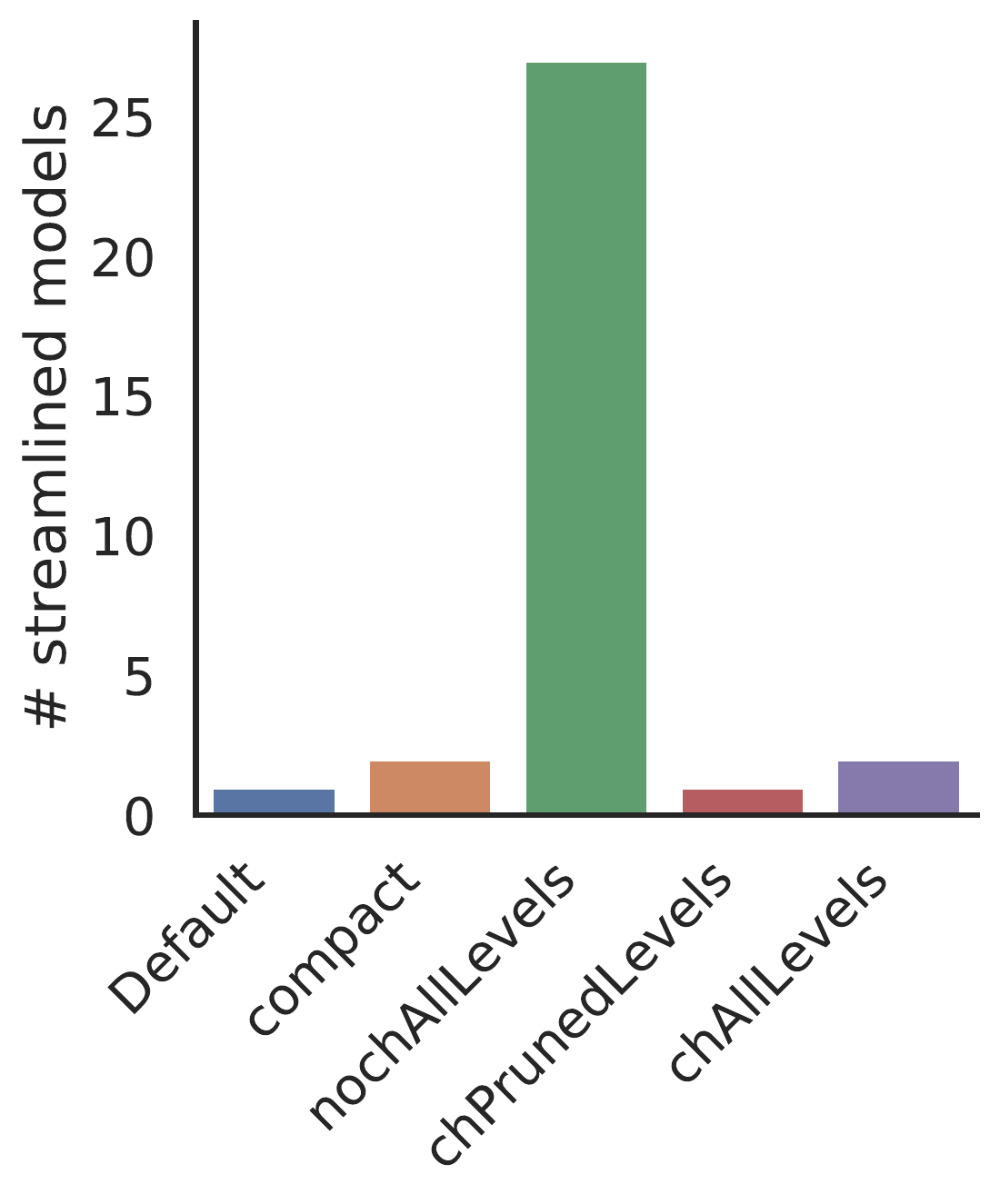}
    \caption{Composition of the portfolio generated by our multi-model search for BACP. The x-axis represents heuristic rules (Table \ref{table:results}) used by \conjure{}. The y-axis shows the number of streamlined models in the portfolio generated by the corresponding heuristics rules. Note that models refined according to the same set of rules can still differ due to the additional modelling choices introduced by the added streamliners.}
    \label{fig:heuristic-count}
\end{figure}

\Cref{fig:heuristic-count} presents the composition of the portfolio of models produced for BACP. It is immediately apparent that \conjure{}'s \default heuristic, which is the only one employed in previous work~\cite{spracklen2018automatic}, is not the one mostly used by the portfolio. In what follows we discuss the composition the portfolio in more details and its connection to the performance improvement observed.

The generated portfolio is comprised of various models generated from five different heuristics. Some models have both \emph{channelling} and \emph{precedence levels} for representations enabled whereas others have both disabled. 
 Those modelling choices interact with the candidate streamliners and enable efficient propagation during the search, leading to the improvement in all performance measurements as shown in Table \ref{table:results_new}.

To give a concrete example Streamliner 1 in BACP enforces that the \emph{curr} function is monotonically increasing. For this streamliner the best model enables \emph{channelling}. The \default{} model chooses only one representation for the \emph{curr} function \texttt{matrix indexed by [int(1..n\_courses)] of int(1..n\_periods)} whereas in the dominating model a second representation is also used \texttt{matrix indexed by [int(1..n\_courses), int(1..n\_periods)] of bool} and \conjure{} will automatically add channelling constraints between these two representations. Having these two representations present in the model makes a large difference in the reduction achieved by Streamliner 1. Another example is for Streamliner 10 which restricts half of the values of the range of the \emph{curr} function to be even. In this case, representing the function as a 2-dimensional Boolean matrix (as above) instead of a 1-dimensional integer matrix allows the streamliner to again achieve a higher overall average reduction.

\section{Conclusion}
We have demonstrated that searching for the right model to pair with a streamliner can improve performance significantly over a fixed, default model using a case study with the Balanced Academic Curriculum Problem. Racing can be used to constrain the computational search cost by removing inferior models early. We have also shown for the first time that Algorithm Selection tools can be used to effectively select streamliners on a per-instance basis. Future work is to extend the experiments to other problems, to increase selection performance to close the gap on the VBS, and to investigate additional modelling choices such as when and how it is beneficial to encode streamliners to different representations such as SAT or SMT.

\paragraph{Acknowledgements} This work is supported by EPSRC grant EP/P015638/1 and used the Cirrus UK National Tier-2 HPC Service at EPCC (\url{http://www.cirrus.ac.uk}) funded by the University of Edinburgh and EPSRC (EP/P020267/1). Nguyen Dang is a Leverhulme Early Career Fellow. 

\bibliographystyle{splncs04}
\bibliography{bib}
\end{document}